# Towards Privacy-Preserving Data-Driven Education: The Potential of Federated Learning

Mohammad Khalil
Centre for the Science of Learning & Technology (SLATE)
University of Bergen
Bergen, Norway
mohammad.khalil@uib.no

Ronas Shakya
Centre for the Science of Learning & Technology (SLATE)
University of Bergen
Bergen, Norway
ronas.shakya@uib.no

Qinyi Liu
Centre for the Science of Learning & Technology (SLATE)
University of Bergen
Bergen, Norway
qinyi.liu@uib.no

*Abstract*— The increasing adoption of data-driven applications in education such as in learning analytics and AI in education has raised significant privacy and data protection concerns. While these challenges have been widely discussed in previous works, there are still limited practical solutions. Federated learning has recently been discoursed as a promising privacy-preserving technique, yet its application in education remains scarce. This paper presents an experimental evaluation of federated learning for educational data prediction, comparing its performance to traditional non-federated approaches. Our findings indicate that federated learning achieves comparable predictive accuracy. Furthermore, under adversarial attacks, federated learning demonstrates greater resilience compared to non-federated settings. We summarise that our results reinforce the value of federated learning as a potential approach for balancing predictive performance and privacy in educational contexts.

## I. INTRODUCTION

Data is central to many data-driven applications across various disciplines. This is evident from the significant technological advancements that have impacted numerous areas of human life. The growing interest in data is particularly notable in discussions about the role of AI in society, especially with the rapid progress of generative AI that requires vast amounts of data to function. At the same time, technological developments have led to an abundance of data generated through human interactions with technology, often referred to as digital data exhaust [3].

In education, as in other disciplines, data-driven applications have been revolutionary. Research institutions and industry partners in education have actively utilized data-driven applications to support teaching and learning in schools, higher education, and workplaces. Examples include distance learning, hybrid learning, digital and virtual learning environments, smart classrooms, and mobile learning. This has been translated through, for example, extracting key insights from students, assessing learning designs, evaluating assessments, providing predictions, and enabling or refining personalization [8; 10]. The works of data-driven applications in education are usually incorporated under the umbrella fields of educational data science, including the subfields of AI in education, educational data mining, and learning analytics.

However, collecting and utilizing data in education, particularly data related to students, raises substantial privacy and ethical concerns. These challenges are further compounded by the vast volume of data involved in educational data science, the diversity of data models employed, and the evolving requirements of data protection legislation [11].

Yet, there are promising attempts to develop solutions, even on a smaller scale, that demonstrate encouraging results. One such approach is the relatively emerging field of federated learning. Federated learning, a decentralized method introduced by Google [4], enables the training of machine learning models by distributing datasets across multiple devices. Instead of transferring entire datasets to a central model, federated learning focuses on training local models on distributed data, preserving privacy by minimizing data movement.

This approach has garnered significant attention from security experts and has shown promising results in other disciplines [5; 17]. Still, its application in education remains limited in the number of studies and practical implementations to date [2].

### A. Purpose of the study

This paper explores the critical role of data in education through the lens of learning analytics where predictive applications are central to shaping and informing intervention strategies to learners. Our research study centres on federated learning, with a particular emphasis on its applicability to classification tasks. Specifically, we aim to:

1. Evaluate the performance of machine learning classification models in predicting key educational outcomes, such as student grades and dropout status, under both federated and non-federated learning paradigms.

2. Examine the robustness of machine learning classification models against label-flipping attacks and compare their performance under federated and non-federated settings.

## II. BACKGROUND AND PRELIMINARIES

### A. Federated Learning

Federated learning is a privacy-preserving technology focused on the decentralization of data sources and data modelling. It is a learning paradigm aimed at addressing the challenge of data silos while ensuring data privacy and security [6]. In a typical federated learning setup, a global model is shared with local devices or nodes, and each device uses its own local data to train that model [6]. Rather than uploading raw data to a central server, only model updates (e.g., parameter gradients) are sent back to the server, which then aggregates them to refine the global model. The key advantage of federated learning lies in its ability to train models without centralizing data or requiring data sharing across devices. Instead, it enables model generation using distributed training data, resulting in a more privacy-conscious approach.

### B. Label flipping attack

A label flipping attack is a type of data poisoning attack in machine learning where adversaries maliciously manipulate training data by altering the labels of samples [18], for instance, flipping a "cat" label to "dog" in an image classification task. This attack aims to degrade model performance, introduce biases, or compromise decision boundaries [18]. Furthermore, in the federated learning setting, label flipping is one of the most significant threats that requires further exploration [9]. Malicious clients can tamper with their dataset labels locally before uploading model updates, subtly compromising the integrity of the global model. Unlike direct data breaches, label flipping is a stealthy attack that often goes undetected due to the decentralized nature of federated learning, making it a prevalent and challenging security risk in distributed learning frameworks.

### C. Federated Learning and Label flipping attack in educational settings

In learning analytics, AI in Education, and broader educational data science, data privacy and protection have traditionally been addressed through anonymization techniques [11]. More recently, innovative approaches such as synthetic data generation [7; 13; 14] and differential privacy [12] have been applied and delivered notable advancements in safeguarding sensitive information. While these methods have shown considerable success, this paper shifts the focus to federated learning that remains underexplored in educational settings. A small but growing number of studies have begun investigating the use of federated learning in learning analytics and educational data mining [2] which provides an additional privacy-preserving option for securing educational data and their modelling. Compared to other methods, federated learning enables collaborative model development without centralizing raw data, thus offering a distinct advantage for maintaining data privacy and security within learning analytics.

Educational datasets often include student performance metrics, engagement levels, or behavioural patterns, with labels such as "at-risk" or "proficient." A label flipping attack in this domain might mislabel students' outcomes (e.g., swapping "fail" with "pass") to distort predictive models used for interventions or resource allocation. Testing label flipping attacks on educational data is critical for three reasons. First, education-focused models demand high fairness and reliability, as errors could disproportionately affect vulnerable groups [14; 15]. Simulating label flipping is important to evaluate models against intentional misguidance in ethically sensitive scenarios. Second, educational datasets frequently exhibit inherent class imbalance (e.g., fewer "at-risk" examples), which amplifies the impact of label manipulation [15]. Third, from an adversary's perspective, educational datasets may offer a more attractive target compared to other attacks like Membership Inference Attacks (MIA). If an adversary's objective is to disrupt predictions for personal or organizational gain, such as manipulating outcomes to influence funding decisions or competitive positioning, the incentive to conduct a label flipping attack is significantly heightened. By stress-testing models under controlled label-flipping conditions, researchers can identify vulnerabilities and develop defences tailored to preserve equitable outcomes in federated learning-driven educational systems.

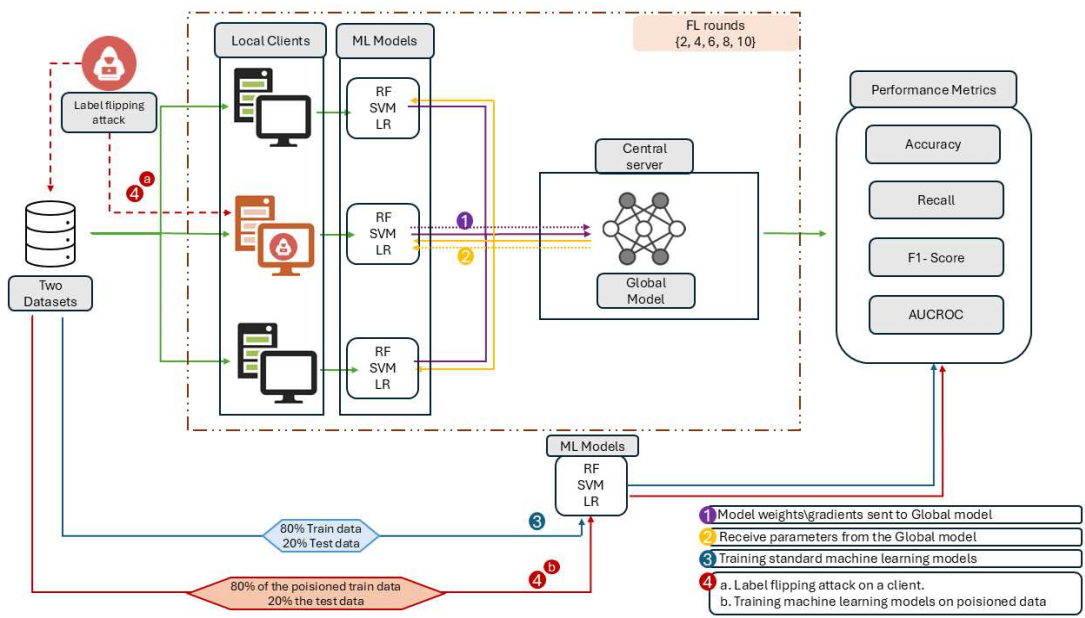

Fig. 1. Pipeline of the study experiement, Federated Learning with Label Flipping Attack and Evaluation Metrics. RF- Random forest, SVM - Support vector machine, LR - Logistic regression, FL - Federated learning.

TABLE I. DESCRIPTION OF THE DATASETS SELECTED FROM THE EDUCATION DOMAIN AFTER PREPROCESSING. THE TARGET VARIABLE IS USED FOR MACHINE LEARNING CLASSIFICATION

| ID | Dataset | Features | #Records | Target Variable | Continuous variable | Categorical variable |
|----|---------|----------|----------|-----------------|---------------------|----------------------|
| A | Student Performance[1] | 33 | 395 | Final_grades | 16 | 17 |
| B | Predict Students' Dropout and Academic Success[2] | 36 | 4424 | Drop_out | 35 | 1 |

[1] https://archive.ics.uci.edu/dataset/320/student+performance - (CC BY 4.0)
[2] https://archive.ics.uci.edu/dataset/697/predict+students+dropout+and+academic+success - (CC BY 4.0)

## III. METHODOLOGY

To evaluate the performance of both federated and non-federated machine learning settings and evaluate their robustness against label-flipping attacks, we developed an overarching pipeline outlining the evaluation process, as shown in Fig. 1.

### A. Dataset description

We selected two tabular datasets for our analysis of educational data (see Table I). The datasets were selected with different sizes. We aim to capture diverse patterns and trends within the education sector.

Dataset A (Student performance) - Includes data on student achievement in secondary education. The features in the dataset include student grades, demographic, social and school related information. This is a questionnaire-based dataset. Originally the dataset is divided into two subjects: Mathematics and Portuguese language. For our experiment, we have used only the mathematics dataset which contains 396 rows and 33 features. Target variable is binary.

Dataset B (Predict Students' Dropout and Academic Success) - This dataset contains information about students enrolled in different education degrees, such as agronomy, design, education, nursing, journalism, management, social service, and technologies with their socio-economic factors, academic path, and demographics. The dataset consists of 4424 rows with 36 features. Target variable is multiclass.

### B. Federated Learning setting

Fig.1 illustrates the pipeline of the federated learning (FL) process used in the experimentation, including the impact of a label flipping attack and the evaluation metrics on federated learning system and non-federated learning. The system consists of three local clients, each training machine learning models on their local datasets. Each client trains its model using 80% of its local data for training and 20% for testing. The models are trained independently on each client without sharing raw data. The clients operate in multiple FL rounds (2, 4, 6, 8, 10), where they send model updates (weights/gradients) to a central server. The server aggregates these updates to form a global model. The global model is then distributed back to the clients for further training. The global model is the result of aggregating updates from all clients. It represents the collaborative learning process across all clients.

### C. Label flipping attack

We performed a random label flipping attack on one of the clients to assess the impact of data poisoning in a federated learning system. In this approach, labels are flipped randomly without targeting a specific class. Around 50 % of the training data were flipped. The compromised client trains its local model on the poisoned data (80% poisoned training data and 20% testing data).

### D. Evaluation of performance

In our experiment, we selected metrics to evaluate the robustness of the federated learning system without label flipping attack and under label flipping attack.

The performance of the federated learning system is also compared with a non-federated learning approach, where models are trained on both standard and poisoned datasets. The evaluation metrics used include accuracy, recall, F1-score, and AUC-ROC. The values are averaged across FL rounds.

**Accuracy** is the proportion of all classifications that were correctly made by a model. It is mathematically defined as:

$$\text{Accuracy} = \frac{correct\ classification}{Total\ classification} \times 100\%$$

represented by percentage (%).

**Recall** is the proportion of all actual positives that were classified correctly as positive. It is mathematically defined as:

$$\text{Recall} = \frac{Correctly\ classified\ actual\ positives}{all\ actual\ positives}$$

the value ranges between 0 and 1. The higher the value, the better are the result.

**F1-Score** is the harmonic mean between recall and precision. It is mathematically represented as:

$$\text{F1-Score} = 2 \times \frac{Precision \times Recall}{Precision + Recall}$$

the value ranges between 0 and 1. The higher the value, the better are the result.

**AUCROC** evaluates model performance across all possible classification thresholds and measures how well a model distinguishes between classes across all thresholds. Its value ranges from 0 to 1: values near 1 indicate strong distinction, around 0.5 reflect random guessing, and near 0 show poor discrimination.

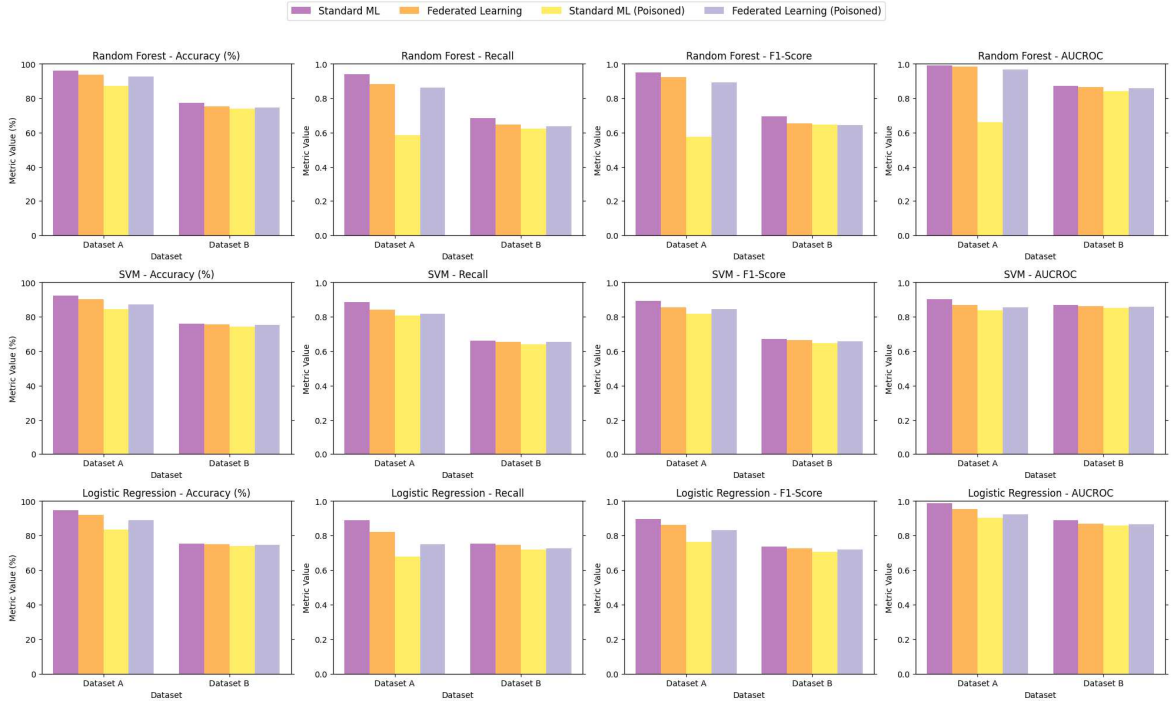

Fig. 2. The average Accuracy, Recall, F1-score, and AUCROC for each dataset and the classifiers used in non-federated learning and federated learning across different rounds. The height of the bar represents the metric values, with higher bars indicating better performance. Best viewed in color.

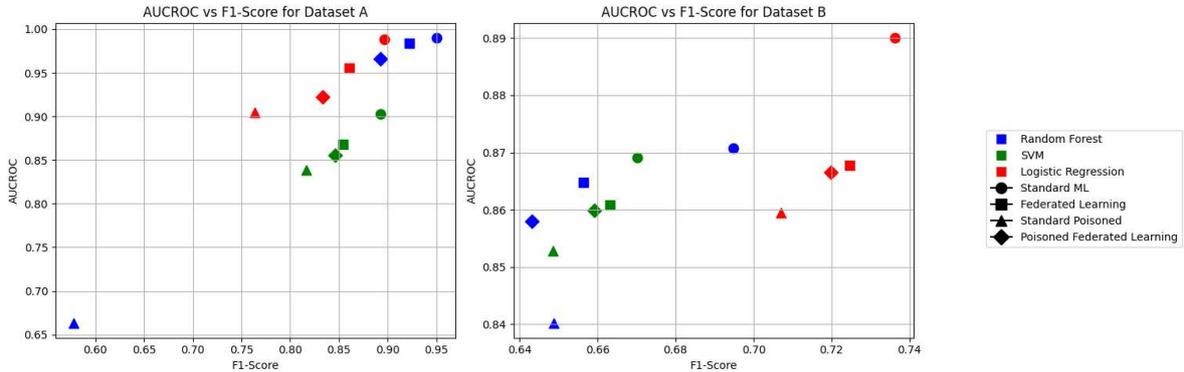

Fig. 3. AUCROC and F1-score for each dataset and the classifiers used in non-federated learning and federated learning. Color represents the models, and shapes represents the methods. Best viewed in color.

## IV. RESULTS

Following the study's pipeline in Fig.1, this section presents our study results[1] of comparing standard machine learning (non-federated) and federated learning for classification tasks on datasets A and B, as well as their performance after a label flipping attack as shown in Fig.2 and Fig. 3.

### A. General Performance of federated and non-federated learning

With respect to the performance of the classification metrics when comparing the federated and non-federated learning, our results indicate a slight decrease in performance across all metrics when comparing federated learning results to standard machine learning (non-federated learning). We noticed that for Dataset A, which is relatively small, the performance degradation is more pronounced than the larger dataset of B. For instance, Logistic Regression in Dataset A experiences a 6.62% drop in Recall while for Dataset B, the difference has been much less significant. This can be explained by the following two factors: the smaller size of Dataset A, and the fact that the target variable in the larger Dataset B is multiclass.

Even though in our experiment we ran a few rounds of communication for federated learning, we observed that as the number of federated learning rounds increases the performance metric values (i.e., Accuracy, Recall, F1-Score, and AUCROC) are affected. For instance, at a certain number of rounds, the values reach a saturation point where there was little to no further improvement. In educational settings, this observation opens up opportunities for future experimentation to determine an optimal range of rounds for maximum performance.

---
[1] *Supporting results are available in the supplementary table at the end of the paper.*

### B. Performance of federated and non-federated learning post the label flipping attack

In general comparison, federated learning showed improved resilience under poisoning conditions. The label-flipping attack on the data impacted model performance, with notable outcomes in federated and non-federated settings. In the non-federated setting, the attack caused a performance drop of up to 30% in the Recall and AUCROC metrics for the Random Forest model in Dataset A. In contrast, the attack's impact on SVM and Logistic Regression models, across both federated and non-federated settings, was slightly negligible.

On the other hand, for Dataset B, we document that the performance results of the machine learning classification with Federated learning have a slight increase in performance but not as much as for Dataset A.

## V. DISCUSSION AND CONCLUSION

This study evaluates the performance of machine learning classification models on educational datasets under both federated and non-federated learning paradigms. The evaluation addressed both a general setting and in the presence of an adversarial scenario called the label-flipping attack. Our findings indicate that federated learning generally performs robustly in both scenarios, with and without the attacks. While a marginal reduction in performance was observed in federated settings, this trade-off is usually justified by enhanced privacy advantages offered through decentralization compared to centralized model training [6].

Moreover, our analysis revealed that the impact of label flipping attacks on the dataset's integrity was with minimal effect as we evidenced by four popular metrics of Accuracy, Recall, F1-Score, and AUCROC. Our results align with [1]'s findings, who highlight the emerging potential of federated learning for machine learning applications within education. This study thus reinforces the value of federated learning as a potential approach for balancing predictive performance and privacy in educational contexts.

In a closer analysis of the impact of the label flipping attack, our experiments indicate that the (random) label flipping attacks tend to have a relatively weak impact compared to other strong attacks such as white box membership interference attack (e.g. [16]) on federated learning models. This observation reveals an interesting gap that requires further attention. In educational applications, where federated learning can be leveraged to predict crucial outcomes like student grades and dropout status, the integrity of the learning process becomes more paramount. The modest effect of random flipping suggests that federated learning systems may be resilient to such naive perturbations in our experiment. However, it also implies that federated learning settings could be vulnerable to more sophisticated, targeted adversarial strategies [16]. In real-world scenarios, attackers might employ stronger, adaptive methods that could significantly compromise federated learning model performance and reliability. Thus, our study highlights a more pressing need to develop and evaluate more strong attacks to fully understand and subsequently develop federated learning frameworks against potential strong and real-world threats in the education domain.

### A. Study limitations

This study has some limitations. First, we focused exclusively on classification models and did not explore the impact of federated learning and poison attacks on other machine learning regression models. Second, our analysis of the classification performance was limited to two datasets, and we only used a small number of federated learning rounds, which may constrain the global model improvements. Finally, we evaluated model performance metrics only and did not investigate privacy metrics or assess the privacy guarantees of the federated learning settings. Future studies can benefit from further exploration of more datasets, more machine learning modelling, and privacy metrics evaluation.

ACKNOWLEDGMENT

This research study was conducted with co-funding from Equinor ASA Akademiaavtalen project "Accelerating Privacy and Data Protection Measures Using Synthetic Data Generation (ASPIRE)".

## Supplementary material

| Dataset | Model | Metrics | Standard ML | FL | Difference | Standard Poisoned | Poisoned FL | Differences |
|---|---|---|---|---|---|---|---|---|
| A | Random forest | Accuracy | 96.2 | 93.64 | 2.56 | 87.34 | 92.79 | 5.45 |
| | | Recall | 0.9412 | 0.8830 | 0.0582 | 0.5877 | 0.8602 | 0.2725 |
| | | F1-Score | 0.9506 | 0.9221 | 0.0285 | 0.5780 | 0.8929 | 0.3149 |
| | | AUCROC | 0.9904 | 0.9840 | 0.0064 | 0.6628 | 0.9660 | 0.3032 |
| | SVM | Accuracy | 92.41 | 90.11 | 2.30 | 84.34 | 87.14 | 2.80 |
| | | Recall | 0.8865 | 0.8417 | 0.0448 | 0.8067 | 0.8184 | 0.0117 |
| | | F1-Score | 0.8929 | 0.8539 | 0.0390 | 0.8163 | 0.8459 | 0.0296 |
| | | AUCROC | 0.9032 | 0.8684 | 0.0348 | 0.8385 | 0.8558 | 0.0173 |
| | Logistic regression | Accuracy | 94.81 | 92.11 | 2.70 | 83.54 | 89.07 | 5.53 |
| | | Recall | 0.8887 | 0.8225 | 0.0662 | 0.6774 | 0.7486 | 0.0712 |
| | | F1-Score | 0.8966 | 0.8605 | 0.0361 | 0.7636 | 0.8329 | 0.0693 |
| | | AUCROC | 0.9886 | 0.9552 | 0.0334 | 0.9042 | 0.9222 | 0.0180 |
| B | Random forest | Accuracy | 77.51 | 75.41 | 2.1 | 73.79 | 74.59 | 0.8 |
| | | Recall | 0.6859 | 0.6492 | 0.0367 | 0.6238 | 0.6389 | 0.0151 |
| | | F1-Score | 0.6948 | 0.6563 | 0.0385 | 0.6488 | 0.6432 | 0.0056 |
| | | AUCROC | 0.8708 | 0.8647 | 0.0061 | 0.8401 | 0.8580 | 0.0179 |
| | SVM | Accuracy | 75.93 | 75.64 | 0.29 | 74.46 | 75.18 | 0.72 |
| | | Recall | 0.6618 | 0.6561 | 0.0057 | 0.6407 | 0.6532 | 0.0125 |
| | | F1-Score | 0.6701 | 0.6632 | 0.0069 | 0.6486 | 0.6592 | 0.0106 |
| | | AUCROC | 0.8691 | 0.8608 | 0.0083 | 0.8528 | 0.8599 | 0.0071 |
| | Logistic regression | Accuracy | 75.25 | 74.93 | 0.32 | 74.01 | 74.75 | 0.74 |
| | | Recall | 0.7525 | 0.7475 | 0.0050 | 0.72010 | 0.7248 | 0.0047 |
| | | F1-Score | 0.7363 | 0.7246 | 0.0117 | 0.70700 | 0.7199 | 0.0129 |
| | | AUCROC | 0.8901 | 0.8678 | 0.0223 | 0.85940 | 0.8665 | 0.0071 |